\documentclass[11pt]{article}

\usepackage[final]{acl}
\usepackage{times}
\usepackage{latexsym}
\usepackage[T1]{fontenc}
\usepackage[utf8]{inputenc}
\usepackage{microtype}
\usepackage{inconsolata}
\usepackage{graphicx}
\usepackage{booktabs}
\usepackage{multirow}
\usepackage{amsmath}
\usepackage{float}
\usepackage{stfloats}
\title{Attribution-Guided Masking for Robust Cross-Domain Sentiment Classification}

\author{
  Shubham Harkare \\
  University of Michigan \\
  \texttt{sharkare@umich.edu} \\\And
  Arvind Yogesh Suresh Babu \\
  University of Michigan \\
  \texttt{savyo@umich.edu} \\\And
  Yash Kulkarni \\
  University of Michigan \\
  \texttt{yashkulk@umich.edu} \\
}

\begin{document}
\maketitle

\begin{abstract}
While pre-trained Transformer models achieve high accuracy on in-domain sentiment classification, they frequently experience severe performance degradation when transferring to out-of-domain data. We hypothesize that this generalization gap is driven by reliance on domain-specific spurious tokens. After demonstrating that post-hoc-token-level attribution drift fails to predict this gap, we propose Attribution-Guided Masking (AGM), a training time intervention that dynamically detects and penalizes highly attributed spurious tokens during fine-tuning. AGM's core component is a gradient based attribution masking loss ($\mathcal{L}_{mask}$), which can optionally be combined with a counterfactual contrastive loss to enforce domain-invariant representations, all without requiring target-domain labels or human annotation. Evaluated in a strict zero-shot transfer setting across four diverse domains with eight random seeds, AGM achieves competitive generalization compared to five strong baselines on the hardest transfer (Sentiment140): $\Delta$ = 0.244 versus DANN (0.264), DRO (0.248), Fish (0.247), and IRM (0.238), while uniquely providing token-level interpretability into which features drive the generalization gap. Our qualitative analysis confirms that AGM suppresses attribution on domain-specific tokens such as @mentions, hashtags, and slang, shifting reliance toward domain-invariant sentiment markers. Our ablation study further confirms that attribution-guided masking is the critical component: removing it or replacing it with random token selection consistently degrades performance on difficult transfers. 
\end{abstract}

\section{Introduction}

Pre-trained Transformer models, such as BERT \citep{devlin2019bert} and RoBERTa \citep{liu2019roberta}, have achieved state-of-the-art performance on in-domain sentiment classification. However, these models frequently suffer performance degradation when applied to out-of-domain data. This generalization gap ($\Delta$) poses a major challenge for deploying natural language processing (NLP) systems in real-world scenarios where target domain distributions are unknown or highly variable.

We hypothesize that this cross-domain failure is driven by models' over-reliance on domain-specific spurious tokens rather than domain-invariant sentiment markers. For example, a model might learn to associate a domain-specific word with positive sentiment simply because it co-occurs with positive labels in the source training data, causing it to fail when that word appears in a different context in the target domain.

To diagnose this vulnerability, we initially explored the Attribution Drift Score (ADS), a novel metric designed to quantify token-level attribution shifts between domains. However, comprehensive testing across multiple formulations revealed no meaningful correlation with the generalization gap (detailed in Section~\ref{sec:analysis}). This negative finding highlights that post-hoc token-level attribution drift is an insufficient signal for predicting transfer difficulty.

Motivated by this limitation, we shift from post-hoc prediction to training-time intervention. We propose Attribution-Guided Masking (AGM), a training framework whose core component an attribution masking loss ($\mathcal{L}_{\text{mask}}$) dynamically detects and penalizes highly attributed, domain-specific spurious tokens during fine-tuning, forcing the model to rely on robust, domain-invariant features. We additionally explore a counterfactual contrastive loss ($\mathcal{L}_{\text{CCL}}$) that can complement masking, though our ablation analysis reveals that $\mathcal{L}_{\text{mask}}$ alone is sufficient and sometimes preferable.

To evaluate the effectiveness of our approach, we compare AGM against five strong domain adaptation and generalization baselines: Domain-Adversarial Training of Neural Networks (DANN) \citep{ganin2016domain}, Invariant Risk Minimization (IRM) \citep{arjovsky2019invariant}, Group Distributionally Robust Optimization (DRO) \citep{sagawa2020distributionally}, and Fish gradient matching \citep{shi2022gradient}. We employ a strict zero-shot transfer protocol across four distinct textual domains (IMDb, Amazon, TripAdvisor, and Sentiment140). Our ablation study confirms that attribution-guided masking is the essential component, with its removal or replacement with random token selection consistently degrading performance on the hardest transfer. This work provides a new pathway for developing sentiment classifiers that rely on intrinsic semantic meaning rather than easily exploitable domain-specific artifacts, while offering token-level interpretability into which features drive the generalization gap.

\section{Data}

To evaluate cross-domain generalization, we utilize four distinct and balanced sentiment classification datasets. Each dataset represents a highly divergent textual domain:
\begin{itemize}
    \item \textbf{IMDb} \citep{maas2011learning}: Long-form, highly structured narrative movie reviews.
    \item \textbf{Amazon} \citep{zhang2015character}: Diverse consumer product reviews.
    \item \textbf{TripAdvisor} \citep{enelpol_booking_2024}: Focused hospitality and hotel reviews.
    \item \textbf{Sentiment140} \citep{go2009twitter}: Noisy, short-form, and informal Twitter data.
\end{itemize}

\paragraph{Dataset Curation Decisions}
Table~\ref{tab:data_stats} details the standardized partitions. We employ a strict ``leave-one-out'' zero-shot transfer protocol: during training, the target domain data is entirely excluded.

\begin{table}[h]
\centering
\small
\begin{tabular}{lr}
\toprule
\textbf{Split} & \textbf{Instances per Domain} \\
\midrule
Train & 10,000 \\
Validation & 2,000 \\
Test & 3,000 \\
ADS (Held-out) & 500 \\
\bottomrule
\end{tabular}
\caption{Standardized dataset statistics across all four domains. The ADS set was carved out prior to splitting to guarantee zero overlap.}
\label{tab:data_stats}
\end{table}

\section{Related Work}

\paragraph{Spurious Correlations in NLP}
Prior work has shown that NLP models frequently exploit dataset artifacts and spurious correlations rather than genuine linguistic patterns \citep{gururangan2018annotation}. This motivates explicit interventions to suppress spurious feature reliance during training.

\paragraph{Domain Adaptation and Generalization}
The vulnerability of large pre-trained language models to out-of-domain distribution shifts has motivated extensive research into unsupervised domain adaptation. Domain-Adversarial Training of Neural Networks (DANN) \citep{ganin2016domain} learns domain-invariant representations via a gradient reversal layer. Invariant Risk Minimization (IRM) \citep{arjovsky2019invariant} seeks predictors that achieve simultaneous optimality across multiple environments. Distributionally Robust Optimization (DRO) \citep{kuhn2024distributionally} provides a general framework for optimizing worst-case performance over a set of distributions. \citet{sagawa2020distributionally} apply group DRO to overparameterized neural networks, demonstrating that it can mitigate reliance on spurious correlations when combined with strong regularization, though it requires group annotations that are often unavailable in cross-domain settings. More recently, gradient-based domain generalization methods such as Fish \citep{shi2022gradient} align inter-domain gradient directions to encourage learning of shared features. While these global representation alignment methods have proven effective on structured data, they often struggle when applied to noisy textual domains where spurious correlations are heavily lexicalized. Our approach differs in operating at the token level rather than the representation level, targeting specific spurious features identified by the model's own attribution behavior.

\paragraph{Feature Attribution and Rationale Extraction}
To interpret the decision-making processes of neural networks, feature attribution methods assign importance scores to input features. Integrated Gradients \citep{sundararajan2017axiomatic} is a widely adopted axiomatic approach that computes the integral of gradients along a path from a baseline to the input, typically used for post-hoc interpretability. Our work leverages a highly efficient Gradient$\times$Input approximation to bring this interpretability directly into the optimization loop.

\paragraph{Explanation-Based Regularization}
\citet{ross2017right} proposed constraining model explanations during training to improve robustness. Their approach requires human-annotated ``right reason'' masks that specify which input features a model should or should not rely on, and focuses on robustness to adversarial inputs rather than cross-domain transfer. In contrast, our proposed Attribution-Guided Masking (AGM) automatically identifies spurious tokens via gradient-based attribution without any external supervision. AGM targets cross-domain zero-shot transfer specifically, and derives its training signal entirely from the model's own attribution behavior, making it scalable to any domain pair without annotation effort.

\section{Methodology}

Our proposed Attribution-Guided Masking (AGM) framework intervenes during the fine-tuning phase to explicitly penalize model reliance on spurious, domain-specific tokens. The training process consists of four primary steps: base classification, attribution computation, counterfactual generation, and contrastive optimization.
\begin{figure}[t]
\centering
\includegraphics[width=\columnwidth]{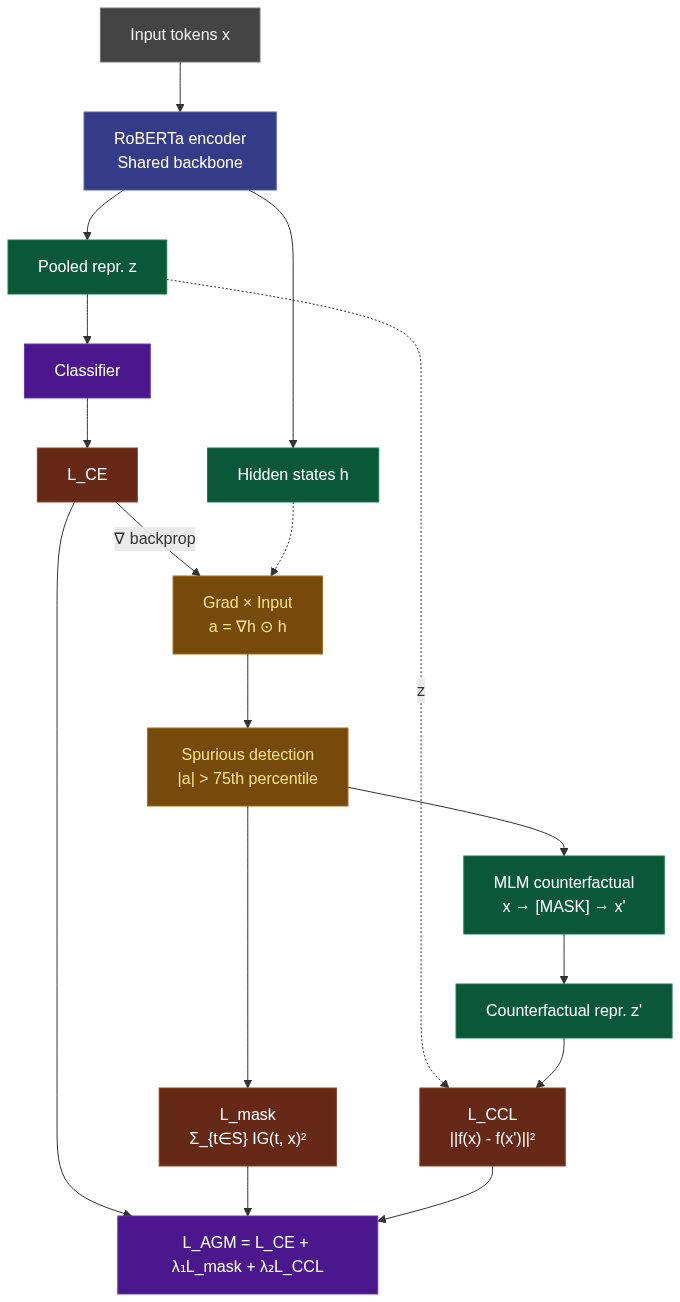}
\caption{AGM training pipeline. Dashed arrow indicates gradient backpropagation for attribution computation.}
\label{fig:architecture}
\end{figure}

\subsection{Base Classification and Attribution}
Given an input sequence of tokens, we first obtain the hidden representations $\mathbf{h}$ from the final layer of a pre-trained RoBERTa \citep{liu2019roberta} encoder. We select RoBERTa-base as our backbone for two reasons: (1) it consistently outperforms BERT on in-domain sentiment classification across all four domains in our baseline experiments (Section~\ref{sec:app_single_source}), making it a stronger starting point for studying cross-domain degradation; and (2) our baseline analysis (Section 5.2) shows that RoBERTa exhibits a \emph{larger} generalization gap than BERT despite its superior in-domain performance, suggesting it more aggressively exploits domain-specific spurious features precisely the behavior AGM is designed to suppress. These representations are pooled to compute the standard cross-entropy classification loss, $\mathcal{L}_{\text{CE}}$.

To identify which tokens the model relies on most, we utilize a Gradient$\times$Input approximation. We compute the gradient of $\mathcal{L}_{\text{CE}}$ with respect to the last hidden state and multiply it by the hidden state itself:
\begin{equation}
    \mathbf{a} = \nabla_{\mathbf{h}} \mathcal{L}_{\text{CE}} \odot \mathbf{h}
\end{equation}
where $\mathbf{a}$ represents the token-level attribution scores. We sum these scores across the hidden dimension to obtain a scalar importance value for each token.

\subsection{Spurious Token Detection and Masking}
Tokens whose absolute attribution magnitude exceeds the $75^{\text{th}}$ percentile ($\tau_{\text{high}} = 0.75$) within a given sequence are flagged as highly attributed and potentially spurious. Let $\mathcal{M}$ denote the set of flagged token indices.

We generate a counterfactual sequence $x'$ by replacing the flagged tokens in the original input $x$ with the \texttt{[MASK]} token. To maintain semantic fluency, we pass this masked sequence through a Masked Language Model (MLM) which shares the same base Transformer weights as our sentiment classifier to predict contextually appropriate replacement tokens. This weight-sharing is not merely a memory-saving optimization; it creates a dynamic, co-adaptive feedback loop in which the classifier's evolving latent space simultaneously shapes the quality of counterfactual replacements. We retain $x'$ only if the base model's predicted label for $x'$ matches the original predicted label for $x$, ensuring the core sentiment remains intact.

\subsection{AGM Training Objective}
To actively discourage the model from relying on spurious tokens, we introduce two auxiliary losses. First, we apply an attribution masking loss, $\mathcal{L}_{\text{mask}}$, to directly penalize high attribution scores on the flagged tokens:
\begin{equation}
    \mathcal{L}_{\text{mask}} = \frac{1}{|\mathcal{M}|} \sum_{i \in \mathcal{M}} (\mathbf{a}_i)^2
\end{equation}

Second, we optionally employ a Counterfactual Contrastive Loss ($\mathcal{L}_{\text{CCL}}$) using Mean Squared Error to align the pooled representation (specifically, the final hidden state of the \texttt{[CLS]} token) of the original input ($\mathbf{z}$) with that of the counterfactual input ($\mathbf{z}'$):
\begin{equation}
    \mathcal{L}_{\text{CCL}} = ||\mathbf{z} - \mathbf{z}'||_2^2
\end{equation}

The full AGM training objective combines these components:
\begin{equation}
    \mathcal{L}_{\text{AGM}} = \mathcal{L}_{\text{CE}} + \lambda_1 \mathcal{L}_{\text{mask}} + \lambda_2 \mathcal{L}_{\text{CCL}}
\end{equation}
where $\lambda_1$ and $\lambda_2$ are hyperparameters weighting the regularization terms (both set to 0.1 in our experiments). As our ablation study demonstrates (Section~\ref{sec:ablation}), $\mathcal{L}_{\text{mask}}$ is the essential component; the mask-only variant ($\lambda_2 = 0$) achieves the strongest results on the hardest transfers.

\section{Evaluation and Results}

\subsection{Experimental Setup}

We evaluate all models using a strict zero-shot, leave-one-out transfer protocol. All models are implemented using the HuggingFace Transformers library \citep{wolf2020transformers} and PyTorch \citep{paszke2019pytorch}. For each fold, one domain is held out as the target and the remaining three serve as source domains. No target-domain data is used during training. For DANN, IRM, DRO, Fish, and AGM, models are trained on the combined source domains; for BERT and RoBERTa, we report the average $\Delta$ across all source--target pairs involving each target. All results are averaged over eight random seeds (42--49) and reported as mean $\pm$ standard deviation.

Our primary metric is the \textbf{Generalization Gap} ($\Delta$):
\begin{equation}
\Delta = |F1_{\text{source}} - F1_{\text{target}}|
\end{equation}
A smaller $\Delta$ indicates a more robust, domain-invariant model. We also report \textbf{Transfer Efficiency} ($TE = F1_{\text{target}} / F1_{\text{source}}$).

\textbf{Hyperparameters.}
All Transformer models use RoBERTa-base with learning rate $2 \times 10^{-5}$, max sequence length 256, warmup ratio 0.1, and early stopping with patience 3. Batch sizes are 32 for baselines and 16 (effective, via gradient accumulation) for AGM. AGM-specific parameters: $\lambda_1 = \lambda_2 = 0.1$, $\tau_{\text{high}} = 0.75$. DANN uses a GRL schedule $\lambda = 2/(1 + \exp(-10p)) - 1$. IRM uses $\lambda = 10^2$ with 500 warmup steps. DRO uses $\eta = 0.01$ with group adjustment $C = 1.5$. Fish uses a Reptile-style inner/outer loop with inner learning rate $1 \times 10^{-4}$.

\textbf{Statistical Significance.}
With eight random seeds, we report 95\% bootstrap confidence intervals (10,000 resamples) for key comparisons. While differences between the strongest methods on Sentiment140 fall within overlapping confidence intervals, we highlight where the intervals are non-overlapping and where results are competitive but not statistically distinguishable.

\subsection{Baseline Analysis: Capacity and Spurious Correlations}

We first establish unadapted baselines using fine-tuned BERT and RoBERTa. A noteworthy pattern emerges: despite achieving stronger in-domain F1, RoBERTa exhibits a \emph{larger} generalization gap than BERT on the hardest transfer. On the Hotel target, RoBERTa's gap widens to $\Delta = 0.060$ compared to BERT's $0.059$. On Sentiment140, the disparity is more pronounced: $0.271$ vs.\ $0.240$.

This pattern is consistent with prior findings that increased model capacity can amplify reliance on spurious correlations during fine-tuning \citep{sagawa2020distributionally, tu2020empirical}. While RoBERTa's richer representations improve in-domain performance, they also enable the model to more aggressively exploit domain-specific shortcuts that fail to transfer.

Full cross-domain transfer matrices detailing the exact single-source to single-target F1 scores for both unadapted BERT and RoBERTa are provided in Appendix~\ref{sec:app_single_source}.

\subsection{Domain Adaptation Baselines}

\textbf{DANN} successfully reduces $\Delta$ to competitive levels on structured review domains ($0.018$--$0.025$ for IMDb, Amazon, Hotel), but suffers catastrophic failure on Sentiment140 ($\Delta = 0.264$). This indicates that global distribution alignment is insufficient when the stylistic distance between source and target is extreme the domain classifier can easily distinguish Twitter text from long-form reviews without learning sentiment-invariant features.

\textbf{IRM} achieves strong results on structured domains ($\Delta = 0.017$--$0.034$) and the lowest mean $\Delta$ on Sentiment140 ($0.238$), but exhibits the highest variance across seeds ($\pm 0.036$). We attribute this instability to the well-documented sensitivity of the IRMv1 penalty to hyperparameter selection \citep{rosenfeld2021risks}; our first configuration ($\lambda = 10^4$, warmup $= 100$) failed entirely, and even the tuned setting ($\lambda = 10^2$, warmup $= 500$) produces inconsistent results across seeds.

\textbf{DRO} minimizes worst-case group loss across source domains, achieving competitive structured domain performance ($\Delta = 0.017$--$0.033$) and Sentiment140 $\Delta = 0.248$. However, it requires explicit group (domain) annotations during training, which limits its applicability in settings where domain boundaries are unknown.

\textbf{Fish} aligns inter-domain gradient directions via a Reptile-style meta-learning loop, achieving the tightest variance among baselines on structured domains. On Sentiment140, Fish achieves $\Delta = 0.247$, comparable to DRO but with lower variance ($\pm 0.021$).

Comprehensive performance metrics for all baselines, including raw Source F1 and Target F1 scores across all transfer pairs, are detailed in Appendix~\ref{sec:appendix_baselines}.

\textbf{Baseline Implementation Details.}
All baseline methods use the same RoBERTa-base encoder as AGM to ensure a fair comparison.

\subsection{AGM Results}

Table~\ref{tab:main_results} presents the main comparison across all models. We report both the full AGM objective and the mask-only variant ($\mathcal{L}_{\text{CE}} + \lambda_1 \mathcal{L}_{\text{mask}}$), since our ablation analysis identifies the latter as the recommended configuration.

The mask-only variant achieves $\Delta$ values of $0.013$ (IMDb), $0.019$ (Amazon), $0.032$ (Hotel), and $0.244$ (Sentiment140). On Sentiment140---the hardest transfer setting---both AGM variants ($\Delta = 0.244$) achieve lower mean $\Delta$ than DANN ($0.264$), DRO ($0.248$), and Fish ($0.247$), while performing comparably to IRM ($0.238$). In absolute terms, AGM achieves Target F1 of $0.706$ on Sentiment140, comparable to Fish ($0.707$), IRM ($0.706$), and DRO ($0.703$), and above DANN ($0.688$). Importantly, AGM achieves these results while uniquely providing token-level interpretability into which features drive the generalization gap. AGM maintains high source F1 ($> 0.91$) across all folds while achieving strong target transfer, confirming that the regularization does not degrade in-domain performance (detailed breakdown in Appendix A, Table~\ref{tab:agm_detailed}).

\textbf{Interpreting the Sentiment140 gap.}
While IRM achieves marginally lower mean $\Delta$ on Sentiment140 ($0.238$ vs.\ AGM's $0.244$), this comparison should be interpreted in context. IRM exhibits substantially higher variance ($\pm 0.036$) compared to AGM ($\pm 0.015$ for the full objective), and required extensive hyperparameter search (an initial configuration failed entirely). The 95\% bootstrap confidence intervals for IRM $[0.218, 0.266]$ and AGM $[0.235, 0.256]$ overlap considerably, indicating the methods are not statistically distinguishable on this metric. For practitioners, AGM's stability and interpretability represent important practical advantages.

\begin{table*}[t]
\centering
\footnotesize
\setlength{\tabcolsep}{4pt}
\begin{tabular}{lcccccccc}
\toprule
\textbf{Target} & \textbf{BERT} & \textbf{RoBERTa} & \textbf{DANN} & \textbf{IRM} & \textbf{DRO} & \textbf{Fish} & \textbf{AGM} & \textbf{Mask-only} \\
\midrule
IMDb    & .119 & .100 & .018 & .024 & .021 & \textbf{.017} & .017 & .013 \\
Amazon  & .055 & .045 & .025 & .034 & .033 & .027 & .021 & \textbf{.019} \\
Hotel   & .059 & .060 & .021 & \textbf{.017} & .017 & .025 & .031 & .032 \\
Sent140 & .240 & .271 & .264 & \textbf{.238} & .248 & .247 & .244 & .244 \\
\bottomrule
\end{tabular}
\caption{Generalization Gap ($\Delta$, lower is better) across all models under the leave-one-out zero-shot transfer protocol. Best result per row in \textbf{bold}. BERT and RoBERTa values are averaged across all source--target pairs; all other methods are mean over 8 seeds.}
\label{tab:main_results}
\end{table*}

\subsection{Ablation Study}
\label{sec:ablation}

To isolate the contribution of each AGM component, we evaluate three ablation variants: (1)~\textbf{No CCL} (mask-only): $\mathcal{L}_{\text{CE}} + \lambda_1 \mathcal{L}_{\text{mask}}$, removing the counterfactual contrastive loss; (2)~\textbf{No Mask}: $\mathcal{L}_{\text{CE}} + \lambda_2 \mathcal{L}_{\text{CCL}}$, removing attribution-based masking and using random token selection for counterfactual generation; (3)~\textbf{Random Mask}: the full objective but with random token selection replacing attribution-guided spurious detection.

\begin{table}[t]
\centering
\footnotesize
\setlength{\tabcolsep}{3pt}
\begin{tabular}{lcccc}
\toprule
\textbf{Target} & \textbf{Full AGM} & \textbf{Mask-only} & \textbf{No Mask} & \textbf{Random} \\
\midrule
IMDb    & .017$\pm$.006 & \textbf{.013$\pm$.001} & .017$\pm$.004 & .016$\pm$.005 \\
Amazon  & .021$\pm$.007 & \textbf{.019$\pm$.004} & .022$\pm$.004 & .018$\pm$.004 \\
Hotel   & .031$\pm$.009 & .032$\pm$.010 & .032$\pm$.009 & .033$\pm$.009 \\
Sent140 & \textbf{.244$\pm$.015} & .244$\pm$.023 & .260$\pm$.028 & .261$\pm$.020 \\
\bottomrule
\end{tabular}
\caption{Ablation study ($\Delta$, lower is better). All values reported as mean $\pm$ std over 8 seeds.}
\label{tab:ablation}
\end{table}

Results in Table~\ref{tab:ablation} reveal two key findings.

\textbf{Finding 1: Attribution-guided masking is the critical component.} The two variants retaining $\mathcal{L}_{\text{mask}}$ with attribution-guided token selection (Full AGM and Mask-only, $\Delta = 0.244$) outperform the two variants without it on the hardest transfer. On Sentiment140, removing masking entirely (No Mask: $\Delta = 0.260 \pm 0.028$) increases the gap by 1.6 points compared to Full AGM ($0.244 \pm 0.015$), and replacing attribution-guided selection with random selection (Random: $\Delta = 0.261 \pm 0.020$) produces a similar degradation. This demonstrates that the model's gradient-based attribution provides a meaningful signal for identifying spurious tokens that random selection cannot replicate.

\textbf{Finding 2: Full AGM and Mask-only achieve equivalent performance.} With 8 seeds, Full AGM and Mask-only are essentially tied on Sentiment140 ($\Delta = 0.244$ vs.\ $0.244$), though Full AGM exhibits tighter variance ($\pm 0.015$ vs.\ $\pm 0.023$). On the structured domains, Mask-only achieves marginally lower $\Delta$ on IMDb and Amazon. This suggests that $\mathcal{L}_{\text{CCL}}$ neither helps nor hurts on average, but may provide a stabilizing effect. For practitioners seeking simplicity, the mask-only configuration ($\mathcal{L}_{\text{CE}} + \lambda_1 \mathcal{L}_{\text{mask}}$) remains the recommended default, as it achieves equivalent results with fewer components and no counterfactual generation overhead.

On the structured review domains (IMDb, Amazon, Hotel), the margins between all variants are small and generally within overlapping standard deviations, indicating that any regularization attribution-guided or otherwise provides sufficient benefit for these relatively easy transfers.

\section{Analysis}
\label{sec:analysis}

\subsection{Attribution Drift Score: A Negative Result}

A central hypothesis of this work was that cross-domain attribution drift could serve as a predictive diagnostic for generalization failure. We defined the Attribution Drift Score as:
\begin{equation}
\text{ADS}(S, T) = 1 - \cos(\overline{IG}_S, \overline{IG}_T)
\end{equation}
where $\overline{IG}_S$ and $\overline{IG}_T$ are mean token-level attribution vectors (computed via Integrated Gradients) using the \emph{source} model on source and target data respectively.

We tested three formulations: \textbf{(1)~Symmetric ADS}, using independent models per domain (Pearson $r = 0.21$); \textbf{(2)~Directional ADS}, using a single source model applied to both domains ($r = 0.007$); \textbf{(3)~Shared-vocabulary ADS}, directional but restricted to tokens appearing in both domains ($r = 0.04$). None produced meaningful correlation with $\Delta$.

The failure of ADS is itself informative. The shared-vocabulary formulation reveals the core issue: tokens shared between domains are predominantly general sentiment words (e.g., \emph{great}, \emph{terrible}, \emph{recommend}) whose attributions are naturally stable across domains. The domain-specific tokens that \emph{cause} spurious reliance precisely the tokens ADS needs to capture are filtered out by the shared-vocabulary requirement. Meanwhile, the directional formulation collapses because a single model produces uniformly high ADS values ($0.72$--$0.96$) across all pairs, lacking the variance needed for correlation.

This negative result motivates the shift from post-hoc prediction to training-time intervention: since we cannot reliably predict which domain transfers will fail, we instead directly suppress spurious token reliance during fine-tuning via AGM.

\subsection{Why DANN Fails on Sentiment140}

DANN achieves strong results on the structured review domains ($\Delta = 0.018$--$0.025$) but suffers catastrophic degradation on Sentiment140 ($\Delta = 0.264$). We attribute this to the nature of DANN's domain alignment mechanism.

When the target domain is stylistically similar to the source (e.g., long-form reviews), the domain classifier must focus on subtle content differences, forcing the feature extractor to learn genuinely transferable sentiment features. However, Sentiment140's Twitter data is so superficially distinct in length, vocabulary, and formality that the domain classifier can achieve high accuracy from surface-level cues alone. The gradient reversal signal then encourages the model to suppress these surface features without learning better sentiment representations, leading to representation collapse.

\subsection{Optimization Tension Between $\mathcal{L}_{\text{mask}}$ and $\mathcal{L}_{\text{CCL}}$}
\label{sec:tension}

Our ablation results show that Mask-only ($\mathcal{L}_{\text{CE}} + \lambda_1 \mathcal{L}_{\text{mask}}$) and Full AGM achieve essentially identical performance on Sentiment140 ($\Delta = 0.244$ vs.\ $0.244$), with Full AGM exhibiting slightly tighter variance ($\pm 0.015$ vs.\ $\pm 0.023$). This suggests that $\mathcal{L}_{\text{CCL}}$ neither consistently helps nor hurts, but may provide a stabilizing effect.

We hypothesize that the two losses introduce competing gradient signals. $\mathcal{L}_{\text{mask}}$ flattens the attribution landscape by reducing attribution on specific tokens, while $\mathcal{L}_{\text{CCL}}$ enforces representational invariance to token replacement---objectives that can conflict when counterfactual replacements are themselves informative tokens. We recommend $\mathcal{L}_{\text{mask}}$ alone as the default configuration due to its simplicity and equivalent performance.

\subsection{Domain Asymmetry in Transfer}

Our results reveal a consistent asymmetry in transfer difficulty. Transferring \emph{to} Sentiment140 is uniformly difficult across all methods (DANN: $0.264$, IRM: $0.238$, DRO: $0.248$, Fish: $0.247$, AGM: $0.244$), while transferring \emph{from} Sentiment140 to cleaner domains is relatively easy. When Sentiment140 is in the training mix and a structured review domain is the target, all models exhibit $TE > 1.0$---target performance \emph{exceeds} source performance (e.g., $TE = 1.023$ for AGM on Amazon). This anomaly is consistent across all methods and seeds, and does not indicate data leakage.

This pattern has a natural explanation. The noise and informality of Twitter data depresses source F1 scores. However, models trained on the diverse source mix are forced to rely on domain-invariant sentiment markers, which then transfer well to cleaner, more structured target domains. The difficulty lies in the reverse direction: models trained on structured reviews learn domain-specific shortcuts that fail catastrophically on Twitter text.

\subsection{Qualitative Token Analysis}

\begin{figure*}[t]
\centering
\includegraphics[width=\textwidth]{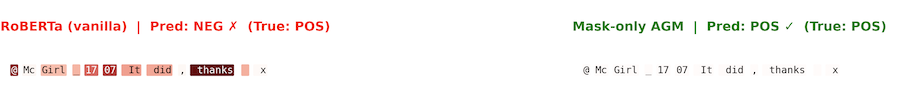}
\caption{Token-level attribution heatmaps on Sentiment140 test examples. Left: vanilla RoBERTa concentrates attribution (red) on domain-specific tokens (@mentions, slang, numbers). Right: Mask-only AGM flattens the attribution landscape, suppressing spurious token reliance. In the bottom example, RoBERTa predicts incorrectly while AGM predicts correctly.}
\label{fig:heatmap}
\end{figure*}

To visualize the effect of attribution-guided masking, Figure~\ref{fig:heatmap} presents token-level attribution heatmaps comparing vanilla RoBERTa and Mask-only AGM on Sentiment140 test examples. Attribution scores are computed via Grad$\times$Input on the final encoder hidden states the same representation space where $\mathcal{L}_{\text{mask}}$ operates during training.

RoBERTa concentrates attribution on domain-specific tokens such as @mentions, numbers, and informal abbreviations, while AGM produces a substantially flattened attribution landscape. In the final example, this shift directly impacts prediction: RoBERTa misclassifies a positive tweet as negative due to spurious token reliance, while AGM predicts correctly. Notably, AGM's spurious token detection operates in a context-dependent manner rather than learning a fixed vocabulary-level blocklist different tokens are flagged in different examples based on the model's attribution behavior for each specific input.

\section{Limitations}

Our study has several limitations. First, we evaluate only on binary English sentiment classification across four domains. The extent to which AGM generalizes to multi-class settings, other languages, or tasks beyond sentiment remains an open question.

Second, AGM introduces additional computational cost from the double backward pass and counterfactual forward pass, resulting in approximately 2--3$\times$ slower training than standard fine-tuning.

Third, while we use eight random seeds, the differences between the strongest methods on Sentiment140 (IRM, DRO, Fish, AGM) fall within overlapping bootstrap confidence intervals. We encourage readers to interpret close comparisons with appropriate caution.

\section{Conclusion}

We investigated the problem of cross-domain generalization in sentiment classification through the lens of token-level attribution. Our work makes three contributions.

First, we demonstrated that post-hoc attribution drift is an unreliable predictor of generalization failure, testing three ADS formulations across 12 transfer pairs and documenting why each fails. This negative result is itself informative: shared tokens between domains carry stable attributions, while the domain-specific tokens responsible for spurious reliance are precisely those excluded from cross-domain comparison.

Second, motivated by this finding, we proposed Attribution-Guided Masking (AGM), a training-time intervention centered on a gradient-based attribution masking loss ($\mathcal{L}_{\text{mask}}$) that dynamically detects and penalizes spurious token reliance. AGM requires no target-domain labels, no human annotation of spurious features, and no domain classifier making it applicable to any zero-shot transfer setting. Our qualitative analysis confirms that AGM successfully shifts token-level attribution away from domain-specific artifacts (such as @mentions, hashtags, and slang) toward domain-invariant sentiment markers.

Third, through comprehensive evaluation against five strong baselines (DANN, IRM, DRO, Fish) and ablation analysis with eight random seeds, we demonstrated that AGM achieves competitive generalization on the hardest transfer (Sentiment140, $\Delta = 0.244$),  achieving lower mean $\Delta$ than DANN ($0.264$), DRO ($0.248$), and Fish ($0.247$), while performing comparably to IRM ($0.238$) with substantially lower variance. Removing attribution-guided masking or replacing it with random token selection consistently degrades performance, confirming that gradient-based attribution provides a meaningful signal for identifying spurious tokens that random perturbation cannot replicate.

Future work should extend AGM to multi-class and multilingual settings, validate the approach on non-sentiment tasks, and explore adaptive $\lambda$ scheduling to resolve the optimization tension between $\mathcal{L}_{\text{mask}}$ and $\mathcal{L}_{\text{CCL}}$.

\bibliography{custom}

\appendix

\section{Detailed Baseline Metrics}
\label{sec:appendix_baselines}

Because our baselines utilize different training setups, we report their detailed performance metrics in two parts. Section~\ref{sec:app_single_source} details the single-source transfer performance for unadapted BERT and RoBERTa. Section~\ref{sec:app_multi_source} details the multi-source (leave-one-out) performance for the domain adaptation baselines (DANN, IRM, DRO, and Fish).

\subsection{Unadapted Single-Source Baselines}
\label{sec:app_single_source}

For the unadapted BERT and RoBERTa baselines, models were fine-tuned on a single source domain and evaluated on all four domains. Tables~\ref{tab:bert_matrix} and~\ref{tab:roberta_matrix} present the full cross-domain F1 matrices. The diagonal represents in-domain (source) F1, while the off-diagonals represent zero-shot target F1. All values are mean over 8 seeds.

\begin{table}[H]
\centering
\footnotesize
\setlength{\tabcolsep}{4pt}
\begin{tabular}{lcccc}
\toprule
\textbf{BERT} & \multicolumn{4}{c}{\textbf{Evaluated Target Domain}} \\
\cmidrule(lr){2-5}
\textbf{Trained On} & \textbf{IMDb} & \textbf{Amazon} & \textbf{Hotel} & \textbf{Sent140} \\
\midrule
IMDb    & (\textit{0.9132}) & 0.9130 & 0.8411 & 0.6901 \\
Amazon  & 0.8816 & (\textit{0.9382}) & 0.9037 & 0.7201 \\
Hotel   & 0.7297 & 0.8355 & (\textit{0.9629}) & 0.6856 \\
Sent140 & 0.7600 & 0.8205 & 0.8947 & (\textit{0.8259}) \\
\bottomrule
\end{tabular}
\caption{Cross-domain F1 matrix for BERT (8 seeds). In-domain performance is bracketed on the diagonal.}
\label{tab:bert_matrix}
\end{table}

\begin{table}[H]
\centering
\footnotesize
\setlength{\tabcolsep}{4pt}
\begin{tabular}{lcccc}
\toprule
\textbf{RoBERTa} & \multicolumn{4}{c}{\textbf{Evaluated Target Domain}} \\
\cmidrule(lr){2-5}
\textbf{Trained On} & \textbf{IMDb} & \textbf{Amazon} & \textbf{Hotel} & \textbf{Sent140} \\
\midrule
IMDb    & (\textit{0.9328}) & 0.9376 & 0.8287 & 0.6223 \\
Amazon  & 0.9094 & (\textit{0.9539}) & 0.9049 & 0.7104 \\
Hotel   & 0.7493 & 0.8732 & (\textit{0.9668}) & 0.7089 \\
Sent140 & 0.8185 & 0.8898 & 0.8811 & (\textit{0.8555}) \\
\bottomrule
\end{tabular}
\caption{Cross-domain F1 matrix for RoBERTa (8 seeds). In-domain performance is bracketed on the diagonal.}
\label{tab:roberta_matrix}
\end{table}

\subsection{Domain Adaptation Baselines (Leave-One-Out)}
\label{sec:app_multi_source}

For DANN, IRM, DRO, and Fish, we utilized the same strict leave-one-out protocol as our AGM models. Models were trained on a combined mix of three source domains and evaluated on the held-out target domain. Table~\ref{tab:baseline_raw} reports the raw Source F1 and Target F1.

\begin{table}[H]
\centering
\footnotesize
\setlength{\tabcolsep}{4pt}
\begin{tabular}{llcc}
\toprule
\textbf{Target} & \textbf{Model} & \textbf{Source F1} & \textbf{Target F1} \\
\midrule
\multirow{4}{*}{IMDb}
 & DANN & 0.9214$\pm$0.0013 & 0.9034$\pm$0.0085 \\
 & IRM  & 0.9048$\pm$0.0080 & 0.8812$\pm$0.0229 \\
 & DRO  & 0.9169$\pm$0.0025 & 0.8958$\pm$0.0087 \\
 & Fish & 0.9237$\pm$0.0022 & 0.9064$\pm$0.0038 \\
\midrule
\multirow{4}{*}{Amazon}
 & DANN & 0.9165$\pm$0.0021 & 0.9411$\pm$0.0044 \\
 & IRM  & 0.9024$\pm$0.0014 & 0.9359$\pm$0.0054 \\
 & DRO  & 0.9113$\pm$0.0026 & 0.9444$\pm$0.0033 \\
 & Fish & 0.9170$\pm$0.0011 & 0.9441$\pm$0.0037 \\
\midrule
\multirow{4}{*}{Hotel}
 & DANN & 0.9143$\pm$0.0009 & 0.8932$\pm$0.0124 \\
 & IRM  & 0.8983$\pm$0.0069 & 0.9091$\pm$0.0167 \\
 & DRO  & 0.9123$\pm$0.0025 & 0.8953$\pm$0.0111 \\
 & Fish & 0.9168$\pm$0.0012 & 0.8920$\pm$0.0112 \\
\midrule
\multirow{4}{*}{Sent140}
 & DANN & 0.9518$\pm$0.0014 & 0.6877$\pm$0.0356 \\
 & IRM  & 0.9445$\pm$0.0040 & 0.7064$\pm$0.0340 \\
 & DRO  & 0.9508$\pm$0.0020 & 0.7028$\pm$0.0290 \\
 & Fish & 0.9542$\pm$0.0018 & 0.7071$\pm$0.0205 \\
\bottomrule
\end{tabular}
\caption{Raw Source and Target F1 scores for all domain adaptation baselines under the leave-one-out protocol. Values are reported as mean $\pm$ std over 8 seeds.}
\label{tab:baseline_raw}
\end{table}
\begin{table}[H]
\centering
\footnotesize
\setlength{\tabcolsep}{2pt}
\begin{tabular}{lcccc}
\toprule
\textbf{Target} & \textbf{Source F1} & \textbf{Target F1} & $\boldsymbol{\Delta}$ & \textbf{TE} \\
\midrule
IMDb    & .920$\pm$.004 & .903$\pm$.010 & .017$\pm$.006 & .982$\pm$.007 \\
Amazon  & .914$\pm$.003 & .935$\pm$.005 & .021$\pm$.007 & 1.023$\pm$.008 \\
Hotel   & .912$\pm$.002 & .881$\pm$.008 & .031$\pm$.009 & .966$\pm$.010 \\
Sent140 & .950$\pm$.002 & .706$\pm$.015 & .244$\pm$.015 & .743$\pm$.016 \\
\bottomrule
\end{tabular}
\caption{Detailed AGM (full objective) results under the leave-one-out protocol. All values reported as mean $\pm$ std over 8 seeds.}
\label{tab:agm_detailed}
\end{table}
\end{document}